\documentclass[11pt,a4paper]{article}
\usepackage[hyperref]{acl2018}
\usepackage{times}
\usepackage{latexsym}
\usepackage{amsmath}
\usepackage{soul}
\usepackage{graphicx}
\usepackage{subcaption}
\usepackage{multirow}
\usepackage{tikz-dependency}
\aclfinalcopy % Uncomment this line for the final submission
\def\aclpaperid{1588} %  Enter the acl Paper ID here

\title{Simpler but More Accurate Semantic Dependency Parsing}

\author{Timothy Dozat\\
        Stanford University\\
        {\tt tdozat@stanford.edu}\\\And
        Christopher D.\ Manning\\
        Stanford University\\
        {\tt manning@stanford.edu}\\}

\date{}

\newcommand{\citea}[1]{\citeauthor{#1}'s (\citeyear{#1})}
\begin{document}
\maketitle
\begin{abstract}
  While syntactic dependency annotations concentrate on the surface or functional structure of a sentence, semantic dependency annotations aim to capture between-word relationships that are more closely related to the meaning of a sentence, using graph-structured representations. We extend the LSTM-based syntactic parser of \citet{DozatManning2017} to train on and generate these graph structures. The resulting system on its own achieves state-of-the-art performance, beating the previous, substantially more complex state-of-the-art system by $0.6\%{}$ labeled F1. Adding linguistically richer input representations pushes the margin even higher, allowing us to beat it by $1.9\%{}$ labeled F1.
\end{abstract} 

\section{Introduction}
The 2014 SemEval shared task on Broad-Coverage Semantic Dependency Parsing \citep{Oepenetal2014} introduced three new dependency representations that do away with the assumption of strict tree structure in favor of a richer graph-structured representation, allowing them to capture more linguistic information about a sentence. This opens up the possibility of providing more useful information to downstream tasks \citep{Reddyetal2017,Schusteretal2017}, but increases the difficulty of automatically extracting that information, since most previous work on parsing has focused on generating trees.

\begin{figure*}
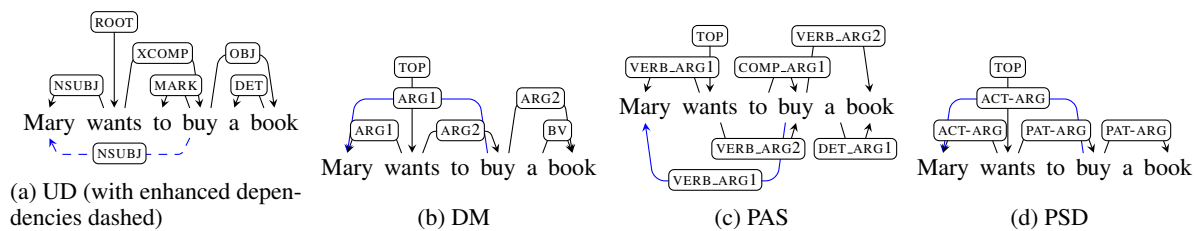

  \small
  \begin{subfigure}[b]{.24\textwidth}
    \centering
    \begin{dependency}[label style={font=\scshape}]
      \begin{deptext}
        Mary \& wants \& to \& buy \& a \& book \\
      \end{deptext}
      \depedge[edge unit distance=2ex]{2}{1}{nsubj}
      \deproot[edge unit distance=2.33ex]{2}{root}
      \depedge[edge unit distance=2ex]{6}{5}{det}
      \depedge[edge unit distance=2.5ex]{2}{4}{xcomp}
      \depedge[edge unit distance=2ex]{4}{3}{mark}
      \depedge[edge unit distance=2.5ex]{4}{6}{obj}
      \depedge[edge below, blue, dashed, edge unit distance=.5ex]{4}{1}{nsubj}
    \end{dependency}
  \caption{UD (with enhanced dependencies dashed)}
  \label{UD-wants}
  \end{subfigure}
  \begin{subfigure}[b]{.24\textwidth}
    \centering
    \begin{dependency}[label style={font=\scshape}]
      \begin{deptext}
        Mary \& wants \& to \& buy \& a \& book \\
      \end{deptext}
      \deproot[edge unit distance=2.33ex]{2}{top}
      \depedge[edge unit distance=1.67ex, blue]{4}{1}{arg1}
      \depedge[edge unit distance=1ex]{2}{4}{arg2}
      \depedge[edge unit distance=2ex]{2}{1}{arg1}
      \depedge[edge unit distance=2.5ex]{4}{6}{arg2}
      \depedge[edge unit distance=2ex]{5}{6}{bv}
    \end{dependency}
    \caption{DM}
    \label{DM-wants}
  \end{subfigure}
  \begin{subfigure}[b]{.24\textwidth}
    \centering
    \begin{dependency}[label style={font=\scshape}]
      \begin{deptext}
        Mary \& wants \& to \& buy \& a \& book \\
      \end{deptext}
      \deproot[edge unit distance=1.65ex]{2}{top}
      \depedge[edge below, edge unit distance=1.8ex, blue]{4}{1}{verb\_arg1}
      \depedge[edge unit distance=2.25ex]{2}{1}{verb\_arg1}
      \depedge[edge unit distance=2.58ex]{4}{6}{verb\_arg2}
      \depedge[edge below, edge unit distance=2.5ex]{5}{6}{det\_arg1}
      \depedge[edge below, edge unit distance=1.25ex]{2}{4}{verb\_arg2}
      \depedge[edge unit distance=2.25ex]{3}{4}{comp\_arg1}
    \end{dependency}
    \caption{PAS}
    \label{PAS-wants}
  \end{subfigure}
  \begin{subfigure}[b]{.24\textwidth}
    \centering
    \begin{dependency}[label style={font=\scshape}]
      \begin{deptext}
        Mary \& wants \& to \& buy \& a \& book \\
      \end{deptext}
      \deproot[edge unit distance=2.33ex]{2}{top}
      \depedge[edge unit distance=1.67ex, blue]{4}{1}{act-arg}
      \depedge[edge unit distance=1ex]{2}{4}{pat-arg}
      \depedge[edge unit distance=2ex]{2}{1}{act-arg}
      \depedge[edge unit distance=1ex]{4}{6}{pat-arg}
    \end{dependency}
    \caption{PSD}
    \label{PSD-wants}
  \end{subfigure}
  \caption{Comparison between syntactic and semantic dependency schemes}
  \label{comparison}
\end{figure*}

Syntactic dependency parsing is arguably the most popular method for automatically extracting the low-level relationships between words in a sentence for use in natural language understanding tasks. However, typical syntactic dependency frameworks are restricted to being tree structures, which limits the number and types of relationships that they can capture. For example, in the sentence \emph{Mary wants to buy a book}, the word \emph{Mary} is the subject of both \emph{want} and \emph{buy}---either or both relationships could be useful in a downstream task, but a tree-structured representation of this sentence (as in Figure \ref{UD-wants}) can only represent one of them.\footnote{Though efforts have been made to address this limitation; see \citet{deMarneffeetal2006,Nivreetal2016,SchusterManning2016,Canditoetal2017} for examples.} 

\citet{DozatManning2017} developed a successful syntactic dependency parsing system with few task-specific sources of complexity. In this paper, we extend that system so that it can train on and produce the graph-structured data of semantic dependency schemes. We also consider straightforward extensions of the system that are likely to increase performance over the straightforward baseline, including giving the system access to lemma embeddings and building in a character-level word embedding model. Finally, we briefly examine some of the design choices of that architecture, in order to assess which components are necessary for achieving the highest accuracy and which have little impact on final performance.

\section{Background}
\subsection{Semantic dependencies}
The 2014 SemEval \citep{Oepenetal2014,Oepenetal2015} shared task introduced three new semantic dependency formalisms, applied to the Penn Treebank (shown in Figure \ref{comparison}, compared to Universal Dependencies \citep{Nivreetal2016}): DELPH-IN MRS, or DM \citep{Flickingeretal2012,OepenLoenning2006}; Predicate-Argument Structures, or PAS \citep{MiyaoTsujii2004}; and Prague Semantic Dependencies, or PSD \citep{Hajicetal2012}.
Whereas syntactic dependencies generally annotate functional relationships between words---such as \emph{subject} and \emph{object}---semantic dependencies aim to reflect semantic relationships---such as \emph{agent} and \emph{patient} (cf.\ semantic role labeling \citep{GildeaJurafsky2002}).
Finally, the SemEval semantic dependency schemes are directed acyclic graphs (DAGs) instead of trees, allowing them to annotate function words as being heads without lengthening paths between content words (as in \ref{DM-wants}).

\subsection{Related work}
Our approach to semantic dependency parsing is primarily inspired by the success of \citet{DozatManning2017} and \citet{DozatQiManning2017} at syntactic dependency parsing and \citet{PengThomsonSmith2017} at semantic dependency parsing. In \citet{DozatManning2017} and \citet{PengThomsonSmith2017}, parsing involves first using a multilayer bidirectional LSTM over word and part-of-speech tag embeddings. Parsing is then done using directly-optimized self-attention over recurrent states to attend to each word's head (or heads), and labeling is done with an analgous multi-class classifier.

\citea{PengThomsonSmith2017} system uses a max-margin classifer on top of a BiLSTM, with the score for each graph coming from several sources. First, it scores each word as either taking dependents or not. For each ordered pair of words, it scores the arc from the first word to the second. Lastly, it scores each possible labeled arc between the two words. The graph that maximizes these scores may not be consistent, with an edge coming from a non-predicate, for example, so they enforce hard constraints in order to prune away invalid semantic graphs. Decisions are not independent, so in order to find the highest-scoring graph that follows these constraints, they use the AD$^3$ decoding algorithm \citep{Martinsetal2011}.

\citea{DozatManning2017} approach to syntactic dependency parsing is similar, but avoids the possibility of generating invalid trees by fully factorizing the system. Rather than summing the scores from multiple modules and then finding the valid structure that maximizes that sum, the system makes parsing and labeling decisions sequentially, choosing the labels for each edge only after the edges in the tree have been finalized by an MST algorithm.

\citet{Wangetal2018} take a different approach in their recent work, using a transition-based parser built on stack-LSTMs \citep{Dyeretal2015}. They extend \citea{ChoiMccallum2013} transition system for producing non-projective trees so that it can produce arbitrary DAGs and they modify the stack-LSTM architecture slightly to make the network more powerful.

\section{Approach}
\subsection{Basic approach}
We can formulate the semantic dependency parsing task as labeling each edge in a directed graph, with \emph{null} being the label given to pairs with no edge between them. Using only one module that labels each edge in this way would be an \emph{unfactorized} approach. We can, however, factorize it into two modules: one that predicts whether or not a directed edge $(w_j, w_i)$ exists between two words, and another that predicts the best label for each potential edge.

\begin{figure}
  \centering%
  \includegraphics[width=\linewidth]{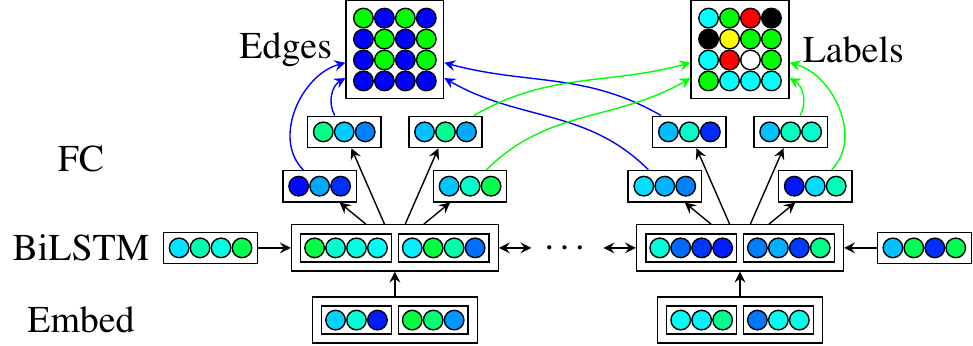}
  \caption{\label{DM17}The basic architecture of our factorized system. Labels are only assigned to word pairs with an edge between them.}
\end{figure}

Our approach closely follows that of \citet{DozatManning2017}. As with many successful recent parsers, we concatenate word and POS tag\footnote{We use the POS tags (and later, lemmas) provided with each dataset.} embeddings, and feed them into a multilayer bidirectional LSTM to get richer representations.\footnote{We follow the convention of representing scalars in lowercase italics $a$, vectors in lowercase bold $\mathbf{a}$, matrices in uppercase italics $A$, and tensors in uppercase bold $\mathbf{A}$. We maintain this convention when indexing and stacking, so $\mathbf{a}_i$ is row $i$ of matrix $A$ and $A$ contains the sequence of vectors $(\mathbf{a}_1, \ldots, \mathbf{a}_n)$.}
\begin{align}
  \mathbf{x}_i &= \mathbf{e}^{(\text{word})}_i \oplus \mathbf{e}^{(\text{tag})}_i\\
  R &= \text{BiLSTM}(X)
\end{align}
For each of the two modules, we use single-layer feedforward networks (FNN) to split the top recurrent states into two parts---a \emph{head} representation, as in Eq.\ (\ref{head1}, \ref{head2}) and a \emph{dependent} representation, as in Eq.\ (\ref{dep1}, \ref{dep2}). This allows us to reduce the recurrent size to avoid overfitting in the classifer without weakening the LSTM's capacity. We can then use bilinear or biaffine classifiers in Eq.\ (\ref{bilin}, \ref{biaff})---which are generalizations of linear classifiers to include multiplicative interactions between two vectors---to predict edges and labels.\footnote{For the labeled parser, $\mathbf{U}$ will be $(d \times c \times d)$-dimensional, where $c$ is the number of labels. For the unlabeled parser, $\mathbf{U}$ will be $(d \times 1 \times d)$-dimensional, so that $s_{i,j}$ will be a single score.}
\newcommand{\eqscaler}[1]{\scalebox{.93}{ \ensuremath{#1}}}
\setlength{\belowdisplayskip}{2pt}%
\setlength{\belowdisplayshortskip}{2pt}%
\begin{align}
  \label{bilin}\eqscaler{\text{Bilin}(\mathbf{x}_1, \mathbf{x}_2)} &\eqscaler{= \mathbf{x}_1^\top\mathbf{U}\mathbf{x}_2}\\
  \label{biaff}\eqscaler{\text{Biaff}(\mathbf{x}_1, \mathbf{x}_2)} &\eqscaler{= \mathbf{x}_1^\top\mathbf{U}\mathbf{x}_2 + W(\mathbf{x}_1\oplus\mathbf{x}_2) + \mathbf{b}}
\end{align}
\setlength{\abovedisplayskip}{2pt}%
\setlength{\abovedisplayshortskip}{2pt}%
\begin{align}
  \label{head1}\mathbf{h}_i^{(\text{edge-head})} &= \text{FNN}^{(\text{edge-head})}(\mathbf{r}_i)\\
  \label{head2}\mathbf{h}_i^{(\text{label-head})} &= \text{FNN}^{(\text{label-head})}(\mathbf{r}_i)\\
  \label{dep1}\mathbf{h}_i^{(\text{edge-dep})} &= \text{FNN}^{(\text{edge-dep})}(\mathbf{r}_i)\\
  \label{dep2}\mathbf{h}_i^{(\text{label-dep})} &= \text{FNN}^{(\text{label-dep})}(\mathbf{r}_i)
\end{align}
\setlength{\belowdisplayskip}{7pt plus2pt minus5pt}%
\setlength{\belowdisplayshortskip}{4pt plus3pt minus3pt}%
\begin{align}
  \eqscaler{s^{(\text{edge})}_{i,j}} &\eqscaler{ = \text{Biaff}^{(\text{edge})}\left(\mathbf{h}_i^{(\text{edge-dep})}, \mathbf{h}_j^{(\text{edge-head})}\right)}\\
  \eqscaler{\mathbf{s}^{(\text{label})}_{i,j}} &\eqscaler{ = \text{Biaff}^{(\text{label})}\left(\mathbf{h}_i^{(\text{label-dep})}, \mathbf{h}_j^{(\text{label-head})}\right)}\\
  y_{i,j}^{\prime(\text{edge})} &= \{s_{i,j} \ge 0\}\\
  y_{i,j}^{\prime(\text{label})} &= \arg\max \mathbf{s}_{i,j}
\end{align}
\setlength{\abovedisplayskip}{7pt plus2pt minus5pt}%
\setlength{\abovedisplayshortskip}{0pt plus3pt}%
The tensor $\mathbf{U}$ can optionally be diagonal (such that $u_{i,k,j} = 0$ wherever $i\not=j$) to conserve parameters. The unlabeled parser (trained with sigmoid cross-entropy) scores every edge between pairs of words in the sentence---these scores can be decoded into a graph by keeping only edges that received a positive score. The labeler (trained with softmax cross-entropy) scores every label for each pair of words, so we simply assign each predicted edge its highest-scoring label. We can train the system by summing the losses, backpropagating error to the labeler only through gold edges. This system is shown graphically in Figure \ref{DM17}.
We find that sometimes the loss for one module overwhelms the loss for the other, causing the system to underfit. Thus we add a tunable interpolation constant $\lambda \in (0,1)$ to even out the two losses.
\begin{align}
  \ell &= \lambda\ell^{(\text{label})} + (1-\lambda)\ell^{(\text{edge})}
\end{align}

Worth noting is that the removal of the maximum spanning tree algorithm and change from softmax cross-entropy to sigmoid cross-entropy in the unlabeled parser represent the only changes needed to allow the original syntactic parser to generate fully graph-structured semantic dependency output. Note also that this system is general enough that it could be used for any graph-structured dependency scheme, including the enhanced dependencies of the Universal Dependencies formalism (which allows cyclic graphs).

\subsection{Augmentations}
\citet{Ballesterosetal2016}, \citet{DozatQiManning2017}, and \citet{Maetal2018} find that character-level word embedding models improve performance for syntactic dependency parsing, so we also want to explore the impact it has on semantic dependency parsing. \citet{DozatQiManning2017} confirm that their syntactic parser performs better with POS tags, which leads us to examine whether word lemmas---another form of low-level lexical information---might also improve dependency parsing performance.

\begin{table*}
  \centering
  \begin{tabular}{lllllllll}
    & \multicolumn{2}{c}{\bf DM} & \multicolumn{2}{c}{\bf PAS} & \multicolumn{2}{c}{\bf PSD} & \multicolumn{2}{c}{\bf Avg}\\
    & ID & OOD & ID & OOD & ID & OOD & ID & OOD\\[1ex]
    \citep{Duetal2015} & 89.1 & 81.8 & 91.3 & 87.2 & 75.7 & 73.3 & 85.3 & 80.8\\
    \citep{AlmeidaMartins2015} & 88.2 &  81.8 & 90.9 &  86.9 & 76.4 & 74.8 & 85.2 & 81.2\\
    WCGL18 & 90.3 & 84.9 & 91.7 & 87.6 & 78.6 & 75.9 & 86.9 & 82.8\\
    PTS17: Basic & 89.4 & 84.5 & 92.2 & 88.3 & 77.6 & 75.3 & 87.4 & 83.6\\
    PTS17: Freda3 & 90.4 & 85.3 & 92.7 & 89.0 & 78.5 & 76.4 & 88.0 & 84.4\\[.5ex]
    Ours: Basic & 91.4 & 86.9 & 93.9 & \bf 90.8 & 79.1 & 77.5 & 88.1 & 85.0\\
    Ours: +Char & 92.7 & 87.8 & \bf 94.0 & 90.6 & 80.5 & 78.6 & 89.1 & 85.7\\
    Ours: +Lemma & 93.3 & 88.8 & 93.9 & 90.5 & 80.3 & 78.7 & 89.1 & 86.0\\
    Ours: +Char +Lemma & \bf 93.7 & \bf 88.9 & 93.9 & 90.6 & \bf 81.0 & \bf 79.4 & \bf 89.5 & \bf 86.3
  \end{tabular}
  \caption{Comparison between our system and the previous state of the art on in-domain (WSJ) and out-of-domain (Brown corpus) data, according to labeled F1 (LF1).}
  \label{performance}
\end{table*}
\section{Results}
\subsection{Hyperparameters\label{hp}}
\begin{table}[h!]
  \centering
  \begin{tabular}{ll}
    \multicolumn{2}{c}{\bf Hidden Sizes}\\
    Word/Glove/POS/ & \multirow{2}{*}{100}\\
    Lemma/Char\\
    GloVe linear & 125\\
    Char LSTM & 1 @ 400\\
    Char linear & 100\\
    BiLSTM & 3 @ 600\\
    Arc/Label & 600\\
    \multicolumn{2}{c}{\bf Dropout Rates (drop prob)}\\
    Word/GloVe/ & \multirow{2}{*}{20\%{}}\\
    POS/Lemma\\
    Char LSTM (FF/recur) & 33\%\\
    Char linear & 33\%\\
    BiLSTM (FF/recur) & 45\%/25\%\\
    Arc/Label & 25\%/33\%\\
    \multicolumn{2}{c}{\bf Loss \&{} Optimizer}\\
    Interpolation ($\lambda$) & .025\\
    $L_2$ regularization & $3e^{-9}$\\
    Learning rate& $1e^{-3}$\\
    Adam $\beta_1$ & 0\\
    Adam $\beta_2$ & .95
  \end{tabular}
  \caption{Final hyperparameter configuration.}
  \label{hyperparams}
\end{table}
We tuned the hyperparameters for our basic system (with no character embeddings or lemmas) fairly extensively on the DM development data. The hyperparameter configuration for our final system is given in Table \ref{hyperparams}. All input embeddings (word, pretrained, POS, etc.) were concatenated. We used 100-dimensional pretrained GloVe embeddings \citep{PenningtonSocherManning2014}, but linearly transformed them to be 125-dimensional. Only words or lemmas that occurred 7 times or more were included in the word and lemma embedding matrix---including less frequent words appeared to facilitate overfitting. Character-level word embeddings were generated using a one-layer unidirectional LSTM that convolved over three character embeddings at a time, whose end state was linearly transformed to be 100-dimensional. The core BiLSTM was three layers deep. The different types of word embeddings---word, GloVe, and character-level---were dropped simultaneously, but independently from POS and lemma embeddings (which were dropped independently from each other). Dropped embeddings were replaced with learned \texttt{<DROP>} tokens. LSTMs used same-mask recurrent dropout \citep{GalGhahramani2016}. The systems were trained with batch sizes of 3000 tokens for up to 75,000 training steps, terminating early after 10,000 steps pass with no improvement in validation accuracy.

\subsection{Performance}
Table \ref{performance} compares our performance with these systems. We use biaffine classifiers, with no nonlinearities, and a diagonal tensor in the label classifier but not the edge classifier. The system trains at a speed of about 300 sequences/second on an Nvidia Titan X and parses about 1,000 sequences/second. \citet{Duetal2015} and \citet{AlmeidaMartins2015} are the systems that won the 2015 shared task (closed track). \emph{PTS17: Basic} represents the single-task versions of \citet{PengThomsonSmith2017}, which they make multitask across the three datasets in Freda3 by adding \emph{frustratingly easy domain adaptation} \citep{DaumeIII2007,KimStratosSarikaya2016} and a third-order decoding mechanism. WCGL18 is \citea{Wangetal2018} transition-based system. Our fully factorized basic system already substantially outperforms \citeauthor{PengThomsonSmith2017}'s single-task baseline and also beats out their much more complex multi-task approach. Simply adding in either a character-level word embedding model (similar to \citea{DozatQiManning2017}) or a lemma embedding matrix likewise improves performance quite a bit, and including both together generally pushes performance even higher. Many infrequent words were excluded from the frequent token embedding matrix, so it makes sense that the system should improve when provided more lexical information that's harder to overfit on.

Surprisingly, the PAS dataset seems not to benefit substantially from lemma or character embeddings.
It has been noted that PAS is the easiest of the three datasets to achieve good performance for; so one possible explanation is that 94\%{} LF1 may simply be near the ceiling of what can be achieved for the dataset. Alternatively, the main difference bewteen PAS as DM/PSD is that PAS includes semantically vacuous function words in its representation. Because function words are extremely frequent, it's possible that they are being disproportionately represented in the loss or LF1 score. Using a hinge loss (like \citet{PengThomsonSmith2017}) instead of a cross-entropy loss might help, since the system would stop focusing on potentially ``easy'' functional predicates once it learned to predict their argument structures confidently, allowing it to put more resources into modeling more challenging phenomena.

\subsection{Variations}
\begin{figure}
  \includegraphics[width=\linewidth]{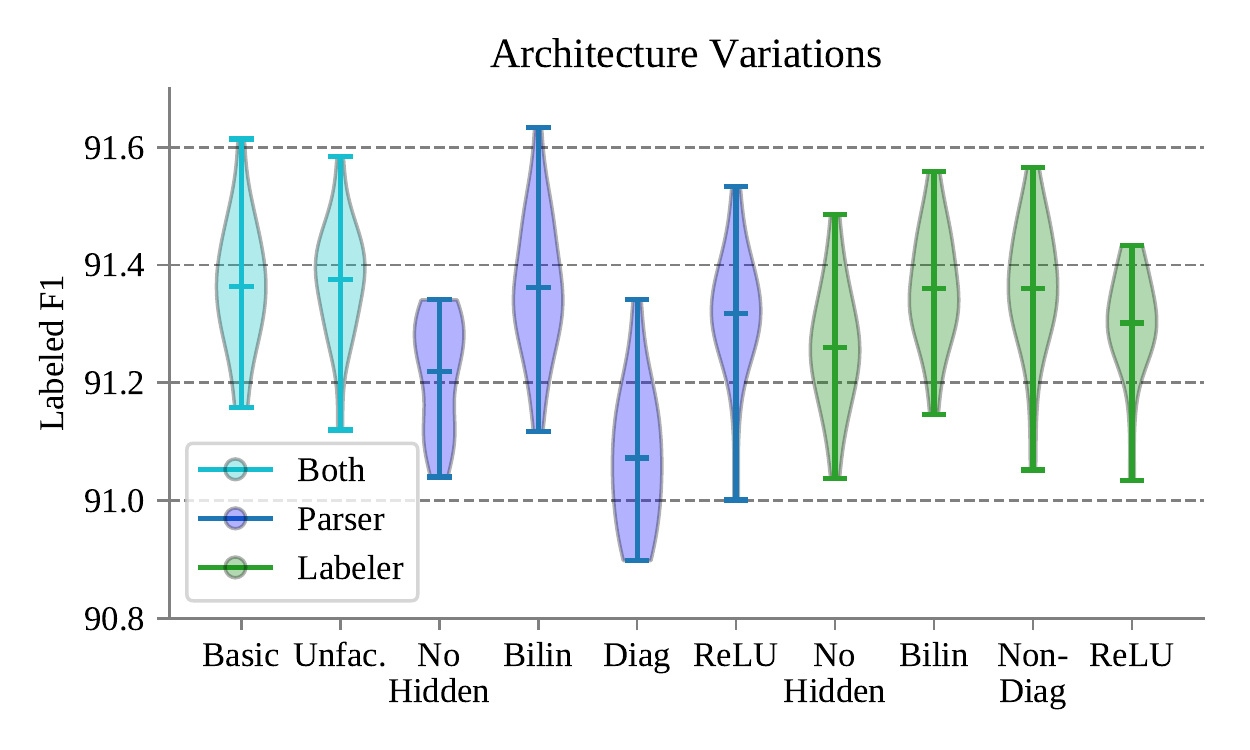}
  \caption{Performance of architecture variations: our basic system; unfactorized (labeler-only); ommitting the hidden layers (Eqs.\ \ref{head1}--\ref{dep2}); with bilinear classifiers (Eq.\ \ref{bilin}); with nondiagonal tensors in the labeler or diagonal tensors in the parser; with the ReLU nonlinearity.}
  \label{variations}
\end{figure}

\thickmuskip=0mu
We also consider the impact that slight variations on the basic architecture have on final performance in Figure \ref{variations}. We train twenty models on the DM treebank for each variation we consider, reducing the number of training steps but keeping all other hyperparameters constant. Rank-sum tests \citep{LehmannDAbrera1975} reveal that the basic system outperforms variants with no hidden layers in the edge classifier $(W=339; p<.001)$ or the label classifier $(W=307; p<.01)$. Using a diagonal tensor $\mathbf{U}$ in the unlabeled parser also significantly hurts performance $(W=388; p<.001)$, likely being too underpowered. While the other variations (especially the unfactorized and ReLU systems) appeared to make a difference during hyperparameter tuning, they were not significant here.

The improved performance of deeper systems (replicating \citet{DozatManning2017}) likely justifies the added complexity. On the other hand, the choice between biaffine and bilinear classifiers comes down largely to aesthetics. This is perhaps unsurprising since the change from biaffine to bilinear represents only a small decrease in overall power. Unusually, using no nonlinearity in the hidden layers in Eqs.\ (\ref{head1}--\ref{dep2}) works as well as ReLU---in fact, using ReLU in the unlabeled parser marginally reduced performance $(W=269; p=.063)$. Overall, the parser displayed considerable invariance to architecture changes. Since our system is significantly larger and more heavily regularized than the systems we compare against, this suggests that unglamorous, low-level hyperparameters---such as hidden sizes and dropout rates---are more critical to system performance than high-level architecture enhancements. 

\section{Discussion}
We minimally extended a simple syntactic dependency parser to produce graph-structured dependencies. Without any further augmentations, our carefully-tuned system achieves state-of-the-art performance, highlighting the importance of finding the best hyperparameter configuration (and by extension, building fast systems that can be trained quickly). Additionally, we can see that a multitask system relying on a complex decoding algorithm to prune away invalid graph structures isn't necessary for achieving the level of parsing performance a simple system can achieve (though it could push performance even higher). We also find easier or independently motivated ways to improve accuracy---taking advantage of provided lemma or subtoken information provides a boost comparable to one found by drastically increasing system complexity. 

Further, we observe a high-performing graph-based parser can be adapted to different types of dependency graphs (projective tree, non-projective tree, directed graph) with only small changes without obviously hurting accuracy. By contrast, transition-based parsers---which were originally designed for parsing projective constituency trees \citep{Nivre2003,AhoUllman1972}---require whole new transition sets or even data structures to generate arbitrary graphs. We feel that this points to graph-based parsers being the most natural way to produce dependency graphs with different structural restrictions.

\bibliographystyle{acl_natbib}
\bibliography{tdozat}

\end{document}